\documentclass{article}
\usepackage{iclr2027_conference,times}   
\usepackage{graphicx}
\usepackage{amsmath, amssymb, amsthm}
\usepackage{mathtools}
\usepackage{booktabs}
\usepackage{xcolor}
\usepackage{url}
\usepackage[colorlinks=true, linkcolor=blue, citecolor=blue, urlcolor=blue]{hyperref}

\newtheorem{proposition}{Proposition}

\theoremstyle{remark}
\newtheorem{remark}{Remark}

\newcommand{\R}{\mathbb{R}}
\newcommand{\E}{\mathbb{E}}
\newcommand{\x}{\mathbf{x}}
\newcommand{\z}{\mathbf{z}}
\newcommand{\mub}{\boldsymbol{\mu}}
\newcommand{\eps}{\boldsymbol{\epsilon}}
\newcommand{\pib}{\boldsymbol{\pi}}
\newcommand{\psib}{\boldsymbol{\psi}}
\newcommand{\gN}{\mathcal{N}}
\newcommand{\Id}{\mathbf{I}}

\DeclareMathOperator{\softmax}{Softmax}
\DeclareMathOperator{\KL}{KL}
\DeclareMathOperator{\Cat}{Cat}

\def\vec#1{{\boldsymbol{#1}}} 
\def\mat#1{{\bf{#1}}} 

\title{Guidance Without Labels: Flow Matching from Coupled Gaussian Mixtures}
\author{Stanislas Ducotterd \qquad Bassam El Rawas \qquad Michael Unser \\
Biomedical Imaging Group, EPFL\\ Lausanne, Switzerland \\
\texttt{\{stanislas.ducotterd, bassam.elrawas, michael.unser\}@epfl.ch}
}
\date{}

\iclrfinalcopy

\begin{document}
\maketitle
\lhead{Preprint}

\begin{abstract}
Flow matching models are typically trained to transport a Gaussian source to the data distribution, with source and target samples paired independently. We replace the source with a fitted Gaussian mixture and couple each data point to a mixture component through an entropic optimal transport problem. We show that the resulting transport reduces to the standard Gaussian-source problem up to a guidance-like correction, which enables the reuse of a pretrained velocity field. This correction is a closed-form function of the fitted mixture plus a term that we learn with a small correction network. We test our model on published checkpoints for CIFAR-10 and CelebA-HQ. It improves FID at every step count, and on CelebA-HQ it reaches the quality of the baseline with about half the steps. The closed-form term alone, with no training, already delivers most of this improvement; the small correction network accounts for the rest.
\end{abstract}

\section{Introduction}
\label{sec:intro}

In generative modeling, one seeks to draw samples from a data distribution $p_1$ available only through a finite set of data points. Flow matching \citep{lipman2023, liu2023rf, albergo2023} achieves this with a velocity field that transports a simple source distribution $p_0$ to $p_1$, learned from the conditional velocity of a single source-data pair. The framework leaves two design choices open, the source $p_0$ itself and the joint distribution with marginals $p_0$ and $p_1$, called the coupling. Standard practice fixes both, taking $p_0 = \gN(\mathbf{0}, \mathbf{I})$ independent of the data distribution. That independence can make sampling inefficient, since every data point is paired with a starting point drawn without regard to where it has to end up. Near the source, a sample carries little information about where it will end up, so the field averages over many modes and the model needs many integration steps, most of them near the noise end where the trajectory decides which mode to commit to \citep{liu2023rf, pooladian2023, luo2026mmfm}. Improving the coupling head-on is difficult. The optimal pairing between a Gaussian and the data is a continuous optimal transport map that can be neither computed nor certified, and minibatch approximations of it are biased toward independence \citep{pooladian2023, tong2024}.

We propose to improve both design choices at once. A source closer to the data shortens the transport, and a coupling that pairs each data point with a nearby source point shortens it further. We propose a Gaussian mixture model (GMM) with $K$ components as the source, $p_0 = \sum_{k=1}^{K} \pi_k\, \mathcal{N}(\mub_k, \tau^2\mathbf{I})$, fitted to the data by maximum likelihood. GMMs have carried density estimation for half a century \citep{dempster1977, mclachlan2000}. They are universal as they can approximate any continuous density to arbitrary accuracy as the number of components $K$ grows \citep{bacharoglou2010, nguyen2020}. Sampling $p_0$ stays easy as it only requires one categorical draw of the component on top of the usual Gaussian one. On top of this, the categorical variable can be coupled to the data perfectly. This allows one to assign a probability distribution over the components for any data point without requiring any optimal-transport-based solvers.

Up to a change of variables and a rescaling of time, the path from a
component of the GMM to the data is the standard Gaussian-source path. A
velocity field pretrained with a Gaussian source can therefore be reused
without any retraining. What the coupling adds is a single term: the gradient
of the log-probability that the noisy image came from the drawn component.
Most of it is a closed-form function of the mixture, and a light network
suffices to learn the rest.

The construction is reminiscent of classifier guidance, in which an externally trained classifier reads a class label off a noisy image and the gradient of its log-probability is added to what a pretrained model predicts \citep{song2021sde, dhariwal2021}. In our approach, it is a consequence of the coupling, and needs no annotation since the components come out of the fit and they supply exact training targets. The granularity of the conditioning follows from $K$ which we are free to choose, as opposed to classifier guidance which is fixed by the number of classes of the dataset. The distribution is left intact as well, since averaging over the component returns $p_1$ exactly, so the coupled model still targets the data distribution. Classifier guidance backpropagates through a classifier at every step, and classifier-free guidance needs no gradients but evaluates the base network a second time \citep{dhariwal2021, ho2022cfg}. Both of these strategies at least double the cost of a step, whereas our correction is a single forward pass through a network far smaller than the base network. Earlier work that starts flow matching from a mixture retrains the velocity network \citep{jia2024structured, issachar2025condprior, luo2026mmfm}. We do not train it at all.

We validate the pipeline on the official Rectified Flow checkpoints \citep{liu2023rf} for CIFAR-10 and CelebA-HQ. The coupled model improves FID at every step count on both datasets, and on CelebA-HQ it reaches with $64$ steps the FID of the baseline at $128$, and with $128$ steps the one at $256$, halving the number of base-network evaluations. Most of this improvement costs nothing: the closed-form part of the correction alone, without any training, already delivers most of the gain and beats the baseline at nearly every step count. The correction network, with about half a percent of the parameters of the base network and under $5\%$ of the cost of a sampling step, adds a further gain. Training it takes $1.4$ hours on CIFAR-10 and $17$ hours on CelebA-HQ on a single GPU, a small fraction of the cost of training the base network.

\section{Related Work}
\label{sec:related}

\paragraph{Flow matching and couplings.}
Flow matching was introduced concurrently by \citet{lipman2023},
\citet{liu2023rf}, and \citet{albergo2023}. Unlike diffusion, where the
forward noising process fixes both the source and its pairing with the data
\citep{ho2020ddpm, song2021sde}, flow matching leaves the coupling free, since
the training loss accepts any joint distribution of source and data with the
right marginals. That freedom has been used for conditional generation,
the source being drawn given the target \citep{albergo2024couplings}, and for
straighter trajectories: minibatch optimal transport
\citep{pooladian2023, tong2024} pairs noise with data inside each batch,
curvature can be penalized directly \citep{lee2023curvature}, and reflow
\citep{liu2023rf} re-pairs noise with the model's own outputs and retrains. All of these act on \emph{continuous}
random variables. Here, by contrast, we act on \emph{discrete} random variables, where optimality is provable and
the solution is closed-form.

\paragraph{Guidance.}
Guidance controls what a generative model produces through an auxiliary
signal, such as a class label or a text prompt. Classifier guidance adds to a
pretrained score or velocity the gradient of an external classifier trained to
predict the signal from the noisy image \citep{song2021sde, dhariwal2021}, and
classifier-free guidance removes the external classifier by training the
generative network with and without the signal and combining the two
predictions \citep{ho2022cfg}. In our case, the signal is simply the mixture
component the sample starts from. Unlike a semantic label or a text prompt, it
requires no annotation, its training targets being closed-form posteriors of
the mixture. Moreover, the simple form of our model yields explicit formulas for the classifier and its gradient, up to a single scalar function.

\paragraph{Mixture sources.}
Gaussian-mixture sources have been proposed before
\citep{jia2024structured,issachar2025condprior,luo2026mmfm}. In all these approaches, the velocity
network is trained from scratch under the new source. We instead view the coupling
through the lens of guidance, under which the conditional velocity is the
unchanged base velocity plus the gradient of a component classifier, a term
that is explicit except for one scalar function. That identity lets
us keep a pretrained base velocity network frozen and shifts the effort to the training of a small
correction network. Different
source mixtures, a sweep over $K$ for instance, can therefore be tried
without retraining the whole pipeline. We also prove that our coupling is the exact entropic-transport
optimum. To our knowledge, ours is the first coupling-side intervention that
works on a pretrained velocity field without touching its weights. 


\section{A Gaussian-Mixture Source and its Optimal Coupling}
\label{sec:gmm-fm}

\subsection{Setup}
\label{sec:setup}

Let $\mat X_0 \sim p_0$ and $\mat X_1 \sim p_1$ on $\R^d$, where $p_1$ is assumed to have a  density $\x_1 \mapsto p_1(\x_1)$ and finite variance. A \emph{coupling} $\Gamma \in \Pi(p_0, p_1)$ is a joint law for the pair $(\mat X_0, \mat X_1)$ whose marginals are $p_0$ and $p_1$, with $\Pi(p_0, p_1)$ denoting the set of all couplings. Flow matching transports the source $p_0$ to the data $p_1$ along $\mat X_t = t \mat X_1 + (1-t) \mat X_0$. The learned object is the velocity field
\begin{equation}
\label{eq:velocity-def}
    \mat v_t(\x) = \E_{(\mat X_0, \mat X_1) \sim \Gamma}\big[\mat X_1 - \mat X_0 \mid \mat X_t = \x\big],
\end{equation}
which generates the marginals, so that integrating $\mat v_t(\x)$ from $\x_0 \sim p_0$ yields $\x_t \sim p_t = \mathrm{Law}(\mat X_t)$ for every $t$. Training parametrizes $\mat v_t$ with a neural network $\mat v_{\theta}(\cdot, t)$ and minimizes the regression $\mathcal{L}(\theta) = \E_{t\sim \mathcal{U}([0,1]),\, (\mat X_0, \mat X_1) \sim \Gamma} \| \mat v_\theta(\mat X_t, t) - (\mat X_1 - \mat X_0)\|^2$, whose minimizer is \eqref{eq:velocity-def}. 

We reserve the letter $\mat u$ for the standard Gaussian-source problem, with state
$\mat Z_t = t \mat X_1 + (1-t)\eps$ for $\eps \sim \gN(\mat 0,\Id)$ independent of $\mat X_1$, marginal $\rho_t$, and velocity $\mat u_t$. Tweedie's formula gives the score-velocity identity, which for a Gaussian source is the one-sided interpolant relation of \citet[Sec. 3.2]{albergo2023unifying}, proved in Appendix~\ref{app:centered} for completeness,
\begin{equation}
\label{eq:centered-tools}
    \mat u_t(\z) = \frac{1}{t}\Big( \z + (1-t)\, \nabla \log \rho_t(\z) \Big).
\end{equation}

\subsection{The entropic coupling is the mixture posterior}
\label{sec:mixture-coupling}

We take as source a Gaussian mixture with $K$ components and a shared isotropic
scale $\tau > 0$,
\begin{equation}
\label{eq:base}
    p_0 = \sum_{k=1}^{K} \pi_k\, \gN(\mub_k, \tau^2\Id), \qquad \mat X_0 = \mub_\kappa + \tau\eps,
\end{equation}
sampled by drawing a component $\kappa \sim \mathrm{Cat}(\pib)$ (i.e., $\Pr(\kappa = k) = \pi_k$) and a noise $\eps \sim \gN(\mathbf{0},\Id)$.

Since $\mat X_0$ is a deterministic function of $(\kappa, \eps)$, choosing the coupling $\Gamma$ amounts to choosing a joint law for the triplet $(\kappa, \eps, \mat X_1)$. If one draws all three independently, nothing links a data point to its starting point and the interpolation lines are free to cross. On the other hand, coupling $\mat X_0$ as a whole to $\mat X_1$ is a continuous optimal transport problem, which is no more tractable for a mixture than for a Gaussian.

Between the two lies the construction of this paper, which couples the
discrete component $\kappa$ to the data and draws the noise $\eps$ independently. A \emph{component coupling} is a coupling $\Gamma \in \Pi(\Cat(\pib), p_1)$ between the component and the data, with $\eps \sim \gN(\mathbf{0},\mathbf{I})$
drawn independently of $(\kappa, \mat X_1)$. Since $\kappa \sim \Cat(\pib)$ and the noise is independent, the map $\mat X_0 = \mub_\kappa + \tau\eps$ gives $\mat X_0 \sim p_0$ exactly, so every component coupling induces an element of $\Pi(p_0, p_1)$.

Two properties make this the productive restriction. It is a $K$-dimensional problem rather than a $d$-dimensional one, solved exactly below (Proposition~\ref{prop:optimal-coupling}) with no transport problem left over. Moreover, because the noise stays independent, each component's conditional path remains a standard Gaussian-source problem, which is what lets a pretrained network supply the velocity $\mat u_t$ (Section~\ref{sec:velocities}).

Among the component couplings we seek the one of least transport cost. Substituting $\mat X_0 = \mub_\kappa + \tau\eps$ and expanding, this cost reads
\begin{align}
\label{eq:cost-split}
    \E\|\mat X_1 - \mat X_0\|^2 &= \E\big\|\mat X_1 - \mub_\kappa - \tau\eps\big\|^2 \nonumber \\
    &= \E \|\mat X_1 - \mub_\kappa\|^2
     - 2\tau\, \E\big\langle \mat X_1 - \mub_\kappa,\, \eps \big\rangle
     + \tau^2\, \E\|\eps\|^2 \nonumber \\
    &= \E \| \mat X_1 - \mub_\kappa\|^2 + \tau^2 d ,
\end{align}
where the cross term vanishes because $\eps$ has mean $\mathbf{0}$ and is independent of $(\kappa, \mat X_1)$, and $\E\|\eps\|^2 = d$. What remains is a \emph{semi-discrete} transport of the atoms $\{\mub_k\}$ with capacities $\pib$ onto $p_1$.

When solved exactly, such a problem returns a hard assignment that sends each data point to a single component. Finding this solution means solving numerically for the $K$ dual variables that enforce the capacities \citep{peyre2019}. We work instead with its \emph{entropic relaxation}, the standard device in computational optimal transport \citep{cuturi2013, peyre2019}. To that end, we add a penalty on how far the assignment strays from the prescribed component frequencies,
\begin{equation}
\label{eq:entropic-ot}
    \min_{\Gamma \in \Pi(\Cat(\pib), p_1)}\; \E_{(\kappa, \mat X_1) \sim \Gamma} \| \mat X_1 - \mub_\kappa\|^2 \;+\; \eta\, \E_{\mat X_1 \sim p_1} \mathrm{KL}\big( \Gamma(\cdot \mid \mat X_1) \,\|\, \Cat(\pib) \big), \quad \eta > 0 .
\end{equation}
The penalty makes the problem strictly convex, so its solution is unique and
available in closed form given the duals.

\begin{proposition}
\label{prop:optimal-coupling}
For any $\eta > 0$, problem \eqref{eq:entropic-ot} has the unique solution
\begin{equation}
    \Gamma_\eta(k \mid \x_1) = \mathrm{Softmax}_k\Big( \log \pi_k + \frac{\psi_k - \|\x_1 - \mub_k\|^2}{\eta} \Big),
\end{equation}
where the duals $\psib \in \R^K$ are determined by the $\Cat(\pib)$-marginal constraint. Moreover, if
\begin{equation}
\label{eq:weight-condition}
    \E_{\mat X_1 \sim p_1}\big[ p_0(k \mid \mat X_1) \big] = \pi_k , \quad k = 1, \dots, K,
\end{equation}
and the temperature is set to $\eta = 2\tau^2$, then the duals vanish, $\psib = \mathbf{0}$, so that the optimal coupling is the posterior responsibility
\begin{equation}
\label{eq:responsibility}
    \Gamma(k \mid \x_1) = p_0(k \mid \x_1) = \mathrm{Softmax}_k\Big( \log \pi_k - \tfrac{\|\x_1 - \mub_k\|^2}{2\tau^2}
    \Big).
\end{equation}
\end{proposition}

The proof of Proposition~\ref{prop:optimal-coupling} is in Appendix~\ref{app:proof-coupling}. When the temperature is set optimally, the optimal coupling is available in closed form with no numerical solve for $\psib$ of the kind semi-discrete transport normally requires \citep{cuturi2013, peyre2019}. Sampling from $\Gamma$ is immediate: draw a training sample $\mat X_1 \sim p_1$, and then $\kappa$ from the closed-form conditional distribution $\Gamma(\cdot \mid \mat X_1)$. Problem \eqref{eq:entropic-ot} is a known reformulation of the mixture likelihood \citep[Theorem 4.2]{rigollet2018}, and \eqref{eq:responsibility} is its solution in the semi-relaxed form \citep[Proposition 2.3]{vayer2025}.

\begin{remark}
\label{rem:shared-scale}
The construction needs the covariance in \eqref{eq:base} to be isotropic and shared. With an anisotropic covariance, the conditional paths are no longer standard Gaussian-source problems, and the pretrained velocity cannot be reused. With one scale $\tau_k$ per component, the posterior weighs the squared distance to $\mub_k$ by $1/(2\tau_k^2)$, and no temperature $\eta$ makes the responsibility optimal for \eqref{eq:entropic-ot}. One shared isotropic $\tau$ matches \eqref{eq:entropic-ot} exactly at $\eta = 2\tau^2$, making the likelihood fit and the transport explicitly solvable in one computation.
\end{remark}

Choosing $\eta=2\tau^2$ eliminates the need to compute the dual variables, provided that Condition~\eqref{eq:weight-condition} holds. Proposition~\ref{prop:stationarity} shows that this condition is automatically satisfied at any converged maximum-likelihood fit with all $K$ mixture weights strictly positive. Its proof is in Appendix~\ref{app:proof-stationarity}.

\begin{proposition}
\label{prop:stationarity}
Fix the means $\{\mub_k\}_{k=1}^{K}$ and the scale $\tau$ of the GMM $p_0$. Consider the expected log-likelihood
\begin{equation}
    L(\pib) = \E_{\mat X_1 \sim p_1} [\log p_0(\mat X_1)]
\end{equation}
as a function of $\pib$ on the open simplex, i.e., $\pi_k > 0$ for all $k=1,\ldots, K$ and $\sum_{k=1}^{K} \pi_k = 1$. Then every stationary point of $L$ under the simplex constraint satisfies Condition \eqref{eq:weight-condition}.
\end{proposition}

\subsection{The coupled velocity is the uncoupled one plus a classifier gradient}
\label{sec:guidance}

Any component coupling partitions the data into \emph{slices}. Since $\kappa \sim \Cat(\pib)$, Bayes' rule on $\Gamma$, the joint law of $(\kappa, \mat X_1)$, gives the law of the data conditioned on the drawn component,
\begin{equation}
\label{eq:slices}
    p_1^{(k)}(\x_1) \coloneqq p_{\mat X_1 \mid \kappa=k}(\x_1) = \frac{\Gamma(k \mid \x_1)\, p_1(\x_1)}{\pi_k}, \qquad \sum_{k=1}^{K} \pi_k\, p_1^{(k)} = p_1.
\end{equation}
Because $\eps$ is independent of $(\kappa, \mat X_1)$, conditioning on the component $\kappa = k$ leaves it standard normal and independent of $\mat X_1$, so the corresponding joint distribution is $p_1^{(k)} \otimes \mathcal{N}(\mathbf{0},\Id)$. In other words, conditioned on $\kappa=k$, the data and the noise are independent. Therefore, conditioned on $\kappa=k$, $\mat Z_t \coloneqq t \mat X_1 + (1-t)\eps$ is an instance of the standard Gaussian-source problem with data law $p_1^{(k)}$, whose marginal we write $\rho^{(k)}_t$ and whose velocity we write $\mat u^{(k)}_t$. Writing $q_t(k \mid \z) \coloneqq \Pr(\kappa = k \mid \mat Z_t = \z)$ for the classifier, Bayes' rule and \eqref{eq:centered-tools} applied to each slice give (Appendix~\ref{app:tilt-derivation})
\begin{equation}
\label{eq:tilt}
    \mat u^{(k)}_t(\z) = \mat u_t(\z) + \frac{1-t}{t} \nabla_{\z} \log q_t(k \mid \z).
\end{equation}
Equation~\eqref{eq:tilt} has the form of classifier guidance \citep{song2021sde, dhariwal2021}, with $\kappa$ in the role of the class label and $q_t(k \mid \z)$ in the role of a classifier reading that label off a noisy image. Two things differ from the usual construction. First, the classifier is internal. Ordinary guidance trains a separate network against annotations supplied from outside the generative model, whereas the target of $q_t$ is a closed-form function of the mixture, as Proposition~\ref{prop:exact-corr} makes explicit. Second, there is no strength parameter. Guidance implementations multiply the gradient by a scale chosen by hand, whereas~\eqref{eq:tilt} is an equality between two exact velocity fields, so $\tfrac{1-t}{t}$ is the only coefficient under which the field generates the slice $p_1^{(k)}$. Mixing over $\kappa$ returns $p_1$, as shown in Proposition~\ref{prop:marginal}, whose proof is in Appendix~\ref{app:proof-marginal}.

\begin{proposition}
\label{prop:marginal}
    Sampling $\kappa \sim \mathrm{Cat}(\pib)$ and $\eps \sim \gN(\mathbf{0}, \Id)$ independently, computing $\x_0 = \mub_\kappa + \tau\eps$, and integrating the exact conditional velocity field $\mat v_t(\cdot \mid k)$ to $t = 1$ produces a sample $\x_1^{(\kappa)} \sim p_1^{(\kappa)}$, and mixing those slice samples over $\kappa$ produces a sample $\x_1 \sim p_1$.
\end{proposition}

\section{From the Guidance Identity to a Trainable Model}
\label{sec:velocities}

A pretrained network supplies the base velocity and the coupling adds the correction of~\eqref{eq:tilt}. We first derive the change of variables that enables the reuse, then isolate the single object the reuse leaves unknown, and then choose how to model it.

\subsection{The Formula of the Velocity}
\label{sec:reparam}

Define
\begin{equation}
\label{eq:clock}
    c(t) = t + (1-t)\tau, \quad s(t) = \frac{t}{t + (1-t)\tau}.
\end{equation}

Proposition~\ref{prop:reparam} shows that, conditioned on a mixture component $\kappa=k$, the GMM interpolation can be expressed in terms of the standard Gaussian-source interpolation, up to a reparameterization of time. Its proof is in Appendix~\ref{app:proof-reparam}.

\begin{proposition}
\label{prop:reparam}
    Under any component coupling, conditioned on $\kappa = k$,
    \begin{equation}
    \label{eq:shift}
        \mat X_t \mid \{\kappa = k\} = c(t) \mat Z_{s(t)} + (1-t)\mub_k ,
    \end{equation}
    and the conditional velocity
    $\mat v_t(\x \mid k) = \E[\mat X_1 - \mat X_0 \mid \mat X_t = \x, \kappa = k]$ is
    \begin{equation}
    \label{eq:cond-velocity-dep}
        \mat v_t(\x \mid k) = \frac{1-\tau}{c(t)}\big(\x - (1-t)\mub_k\big) + \frac{\tau}{c(t)}\, \mat u^{(k)}_{s(t)}\!\left(\frac{\x - (1-t)\mub_k}{c(t)}\right) - \mub_k .
    \end{equation}
    The marginal velocity is the posterior mixture
    \begin{equation}
        \mat v_t(\x) = \sum_{k=1}^{K} p_t(k \mid \x) \mat v_t(\x \mid k), \quad p_t(k \mid \x) \coloneqq \Pr(\kappa = k \mid \mat X_t = \x).
    \end{equation}
\end{proposition}

Conditioned on $\kappa = k$, Equation~\eqref{eq:shift} shows that the mixture
path is the standard Gaussian-source path, up to an affine transformation and
a rescaling of time. The conditional velocity~\eqref{eq:cond-velocity-dep}
therefore only needs the slice velocity $\mat u^{(k)}_{s(t)}$, which
by~\eqref{eq:tilt} is the pretrained velocity plus a classifier gradient, so
the pretrained network is reused as is and only the gradient remains to be
determined. The network is evaluated at the shifted time
$s(t)$, which is exactly the timestep shift of \citet{esser2024sd3} with
shift parameter $1/\tau$, chosen empirically there and derived here.

\subsection{What has to be learned}
\label{sec:whatlearned}

\begin{proposition}
\label{prop:exact-corr}
Let $\sigma_t^2 = t^2\tau^2 + (1-t)^2$ and set
\begin{equation}
\label{eq:mk}
    \mat m_k(\z, t) = \frac{(1-t)^2 \mub_k + t\tau^2 \z}{\sigma_t^2} ,
    \quad
    \nu_t = \frac{\tau^2 (1-t)^2}{\sigma_t^2} ,
    \quad
    F_t(\mat m) = \E_{\mat X \sim \gN(\mat m,\, \nu_t \Id)} \!\left[ \frac{p_1(\mat X)}{p_0(\mat X)} \right] .
\end{equation}
Then the classifier is
\begin{equation}
\label{eq:logits}
    q_t(k \mid \z) = \softmax_k \ell_k(\z, t),
    \quad
    \ell_k(\z, t) = \log \pi_k
    - \frac{\|\z - t\mub_k\|^2}{2\sigma_t^2}
    + \log F_t\big( \mat m_k(\z, t) \big) ,
\end{equation}
and the correction of \eqref{eq:tilt} is
\begin{align}
\label{eq:corr-exact}
    \mat u^{(k)}_t(\z) - \mat u_t(\z)
    ={}&
    \frac{1-t}{\sigma_t^2}\bigg( \mub_k - \sum_{j=1}^{K} q_t(j \mid \z)\, \mub_j \bigg)
    \nonumber\\
    &+
    \frac{(1-t)\tau^2}{\sigma_t^2}
    \bigg( \big[\nabla \log F_t\big](\mat m_k)
    - \sum_{j=1}^{K} q_t(j \mid \z) \big[\nabla \log F_t\big](\mat m_j) \bigg).
\end{align}
\end{proposition}

The proof of Proposition~\ref{prop:exact-corr} is in Appendix~\ref{app:proof-exact-corr}. Equations~\eqref{eq:logits} and~\eqref{eq:corr-exact} are available in closed-form from the mixture parameters, up to the scalar function $F_t$. This is the only remaining quantity that needs to be modeled.


\subsection{Parameterization}
\label{sec:parameterization}

Everything in Proposition~\ref{prop:exact-corr} is explicit except $F_t$, which appears as $\log F_t$ in the logits~\eqref{eq:logits} and as $\nabla \log F_t$ in the correction~\eqref{eq:corr-exact}.  The most direct design would learn the single scalar function $\log F_t$ and compute the gradient from it. We do not, for two reasons. First, computing the gradient at every sampling step would be cost prohibitive. Second, learning a good potential is not learning a good gradient: a regression on the potential controls its values but not its derivatives, and an error that is small yet oscillates fast leaves the potential nearly exact while perturbing the gradient by an arbitrary amount. Accordingly, in energy-based diffusion models, parameterizing the score directly outperforms differentiating a learned energy \citep{salimans2021ebm}. 

For the design, we drop the term $\log F_t$ and represent the second term of~\eqref{eq:corr-exact} with a neural network
\begin{equation}
\label{eq:param}
    \hat{\mat u}^{(k)}_t(\z) = \mat u_t(\z) +
    \frac{1-t}{\sigma_t^2}
    \Big( \mub_k - \sum_{j} q^{\mathrm{ref}}_t(j \mid \z)\, \mub_j \Big)
    +
    \frac{(1-t)\tau^2}{\sigma_t^2}\,
    \mat W_{\phi}\big(\mat m_k,\, \z,\, q^{\mathrm{ref}}_t(\cdot \mid \z),\, t\big),
\end{equation}
where
\begin{equation}
\label{eq:qref}
    q^{\mathrm{ref}}_t(k \mid \z) = \softmax_k \left( \log \pi_k - \frac{\|\z - t\mub_k\|^2}{2\sigma_t^2} \right).
\end{equation}
The mean-squared-error training of $\mat W_{\phi}$ can correct for the missing $\log F_t$ term inside $q_t$, since the error it induces in the first term of~\eqref{eq:param} is absorbed into the regression target of $\mat W_{\phi}$.

The convolutional network $\mat W_{\phi}$ is small, with $(\mat m_k, \z)$ as spatial input and its conditioning vector splits evenly between a sinusoidal embedding of $t$ and a learned map of $q^{\mathrm{ref}}_t(\cdot \mid \z)$. Its architecture and initialization details are in Appendix~\ref{app:details}.

The network $\mat W_{\phi}$ is trained by the flow matching regression with the component drawn from the coupling,
\begin{equation}
\label{eq:head-loss}
    \mathcal{L}(\phi)
    = \E \left\| \hat{\mat u}^{(\kappa)}_t(\mat Z_t) - (\mat X_1 - \eps) \right\|^2 ,
    \quad
    \mat X_1 \sim p_1, \quad
    \kappa \sim \Gamma(\cdot \mid \mat X_1), \quad
    t \sim \mathcal{U}(0, 1),
\end{equation}
with $\mat Z_t = t \mat X_1 + (1-t)\eps$. Its minimizer is the exact slice velocity \eqref{eq:tilt}, which is what lets the training absorb the missing $\log F_t$ term of $q^{\mathrm{ref}}_t$.

Finally, setting $\mat W_{\phi} = \mathbf{0}$ in~\eqref{eq:param} yields a model that requires no training at all, the correction reducing to its closed-form term under $q^{\mathrm{ref}}_t$. We report its results alongside the trained model in Section~\ref{sec:experiments}, under the name ``GMM untrained'', to show that even without training the coupling and $q^{\mathrm{ref}}_t$ already improve generation.

\section{Experiments}
\label{sec:experiments}

Every mixture is fitted by maximizing the log-likelihood
\begin{equation}
\label{eq:gmm_fit}
    \E_{\mat X_1 \sim p_1} \big[ \log p_0(\mat X_1) \big]
\end{equation}
by full-batch gradient descent with Adam, learning the means, the weights
and the single shared scale $\tau$ together (Appendix~\ref{app:recipe}). Sampling uses a uniform Euler grid throughout, in $t$ for the Gaussian
baseline and in $s(t)$ of Equation~\eqref{eq:clock} for every GMM
model.

\subsection{A two-dimensional illustration}
\label{sec:toy}

We run the whole construction for $d=2$, where trajectories can be seen. The target is the checkerboard
density commonly used in the flow matching literature. The base velocity network is an MLP
trained with the standard Gaussian source.
We fit the Gaussian mixture with $K = 64$,
deliberately mismatched with the eight squares of the density so that the
source does not know the mode structure. The correction network is an MLP with $4\%$
of the base velocity network parameters.

\begin{figure}[t]
\centering
\includegraphics[width=0.92\textwidth]{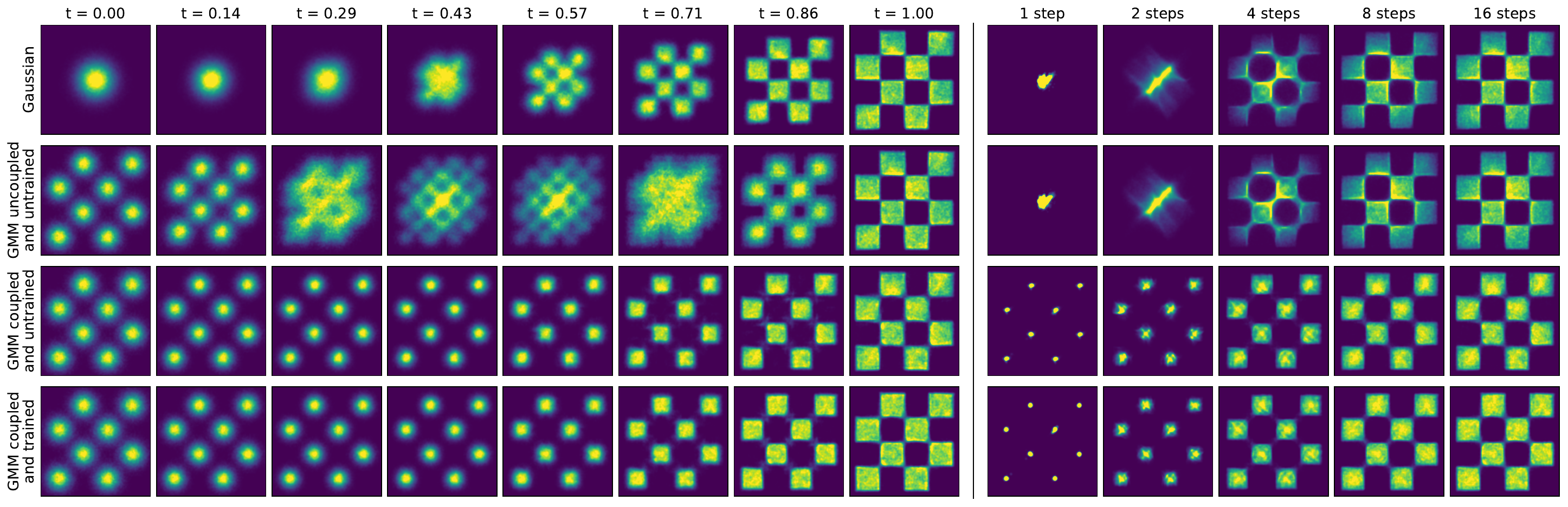}
\caption{Flow matching toward the checkerboard density with different
sources and couplings, one model per row. Left: marginals at intermediate
times $t$ along a $256$-step Euler integration. Right: the final distribution
at $t = 1$ when sampling with a small number of Euler steps.}
\label{fig:checkerboard}
\end{figure}

Figure~\ref{fig:checkerboard} reports two experiments on the same four
models. On the left, every model is integrated with $256$ Euler steps, and
the marginals are shown at intermediate times. On the right, every model is
integrated with a small number of Euler steps and only the final distribution
at $t = 1$ is shown. The Gaussian row is standard flow matching, with the
source drawn independently of the target. The three remaining rows start from
the fitted mixture; although it has $64$ components, maximum likelihood puts
them successfully on the eight squares. The uncoupled row needs no correction, since under the
independent coupling every slice equals $p_1$ and the conditional field of
\eqref{eq:cond-velocity-dep} reduces to the base network. The two coupled rows
add the correction of \eqref{eq:corr-exact}, first with $\mat W_{\phi} = \mathbf{0}$, then with the trained $\mat W_{\phi}$.

The figure disentangles the three ingredients of the construction, the
source, the coupling, and the learned correction, and the four rows show that
most of the improvement comes from the coupling. The source alone changes nothing, the
uncoupled path dissolves the structure it starts with, every component
transporting to every square. It arrives at essentially the same
distribution as the Gaussian model at $t = 1$. The coupling changes
everything. The eight clusters never dissolve, they expand into their
squares. For the coupled models, the checkerboard is already recognizable
from four steps, while the Gaussian and uncoupled models lag at every step count. The trained $\mat W_{\phi}$ improves slightly on the closed form by filling the final
squares more uniformly. The experiment shows that generation is improved just by fitting a GMM and adding a closed-form term to the velocity.

\subsection{Setup of the image experiments}
\label{sec:image-setup}

The image experiments run on the official Rectified Flow checkpoints
\citep{liu2023rf}: DDPM++ on CIFAR-10 ($61.8$M parameters, trained
with horizontal flips) and NCSN++ on CelebA-HQ $256$ ($65.6$M, trained
without). Throughout, each training set is the one the base network was trained on: the CIFAR-10 network saw horizontal flips, so
its training set is the flip-symmetrized one with $100$k images, while the
CelebA-HQ network saw none, so its training set is the $30$k original images. The gradient of \eqref{eq:gmm_fit} is accumulated in
chunks since the whole dataset does not hold in memory.

We sweep $K$ over powers of four and select the model with the highest log-likelihood on the test set of CIFAR-10. It peaks at $K = 4096$ and overfits at $K = 16384$
(Table~\ref{tab:ksweep}), so we keep the former, about $24$ training images per
component. CelebA-HQ has no test split, so we match the average number of training images per
component: $K = 1024$ for $30$k images.  The learned scale is $\tau = 0.313$ on CIFAR-10 and $\tau=0.342$ on CelebA-HQ.

\begin{table}[h]
\centering
\small
\begin{tabular}{lccccc}
\toprule
$K$ & 64 & 256 & 1024 & $\mathbf{4096}$ & 16384 \\
\midrule
train & $-0.412$ & $-0.358$ & $-0.305$ & $-0.259$ & $-0.178$ \\
test  & $-0.412$ & $-0.366$ & $-0.331$ & $\mathbf{-0.319}$ & $-0.330$ \\
\bottomrule
\end{tabular}
\caption{CIFAR-10 log-likelihood per dimension of the fitted mixture, on train and test set.}
\label{tab:ksweep}
\end{table}

The total-variation gap between the two sides of
\eqref{eq:weight-condition} falls below $10^{-4}$,
the hypothesis of Proposition~\ref{prop:optimal-coupling} holds up to numerical accuracy. 

One network $\mat W_{\phi}$ is trained per dataset, with the same architecture
(Appendix~\ref{app:details}) and the same recipe on both
(Appendix~\ref{app:recipe}). The network $\mat W_{\phi}$ has $0.352$M parameters on CIFAR-10
($K = 4096$) and $0.328$M on CelebA-HQ ($K = 1024$), about half a percent of
the base velocity network. Training minimizes \eqref{eq:head-loss}. Its cost is small and
dominated by the frozen base network, which supplies one forward pass per
training sample while receiving no gradient. On one V100, training takes
$1.4$ hours on CIFAR-10 and $17$ hours on CelebA-HQ $256$. Trying a
different source, a new $K$ for instance, therefore costs a mixture fit and
a new $\mat W_{\phi}$, with the base network untouched.

\subsection{Few-step generation}
\label{sec:fewstep}
Our pipeline reproduces the published numbers up to the known discrepancy
between TF-hub and torch Inception features (Appendix~\ref{app:recipe}). We evaluate FID
against the full training set, the published protocol for these checkpoints,
with $50$k generated samples on CIFAR-10 and $10$k on CelebA-HQ, plus the Inception
Score \citep{salimans2016is} on CIFAR-10.

Table~\ref{tab:fid} reports the results on both datasets. The coupled model improves FID at every step count. On CelebA-HQ, its $64$-step FID of $11.24$ already beats the
baseline's $128$-step $11.45$, and its $128$-step $9.80$ the baseline's
$256$-step $10.01$, so the coupling more than halves the number of
base-network evaluations needed to reach either quality. The margin narrows as the
step count grows, which is what the theory predicts, since with exact fields
both models target the same marginal
(Proposition~\ref{prop:marginal}). Appendix~\ref{app:additional} reports the untrained model at every $K$ of the sweep. Larger mixtures help most at small step counts, and the differences between them close as the step count grows.

\begin{table}[h]
\centering
\small
\setlength{\tabcolsep}{3pt}
\begin{tabular}{lccccccccc}
\toprule
Euler steps & 1 & 2 & 4 & 8 & 16 & 32 & 64 & 128 & 256 \\
\midrule
\multicolumn{10}{l}{CelebA-HQ $256$, FID $\downarrow$} \\
\quad Gaussian & 258.40 & 182.43 & 137.16 & 89.01 & 48.53 & 25.08 & 15.00 & 11.45 & 10.01 \\
\quad GMM untrained & 186.94 & 123.86 & 93.61 & 63.60 & 37.12 & 19.91 & 12.81 & 10.63 & 9.63 \\
\quad GMM trained & \textbf{182.97} & \textbf{115.04} & \textbf{87.38} & \textbf{58.90} & \textbf{33.40} & \textbf{16.69} & \textbf{11.24} & \textbf{9.80} & \textbf{9.42} \\
\midrule
\multicolumn{10}{l}{CIFAR-10, FID $\downarrow$} \\
\quad Gaussian & 371.25 & 173.07 & 52.95 & 17.20 & 8.59 & 5.47 & 4.13 & 3.43 & 3.05 \\
\quad GMM untrained & 151.43 & \textbf{85.62} & 31.36 & 14.49 & 8.48 & 5.65 & 4.14 & 3.32 & 2.98 \\
\quad GMM trained & \textbf{151.33} & 87.01 & \textbf{30.52} & \textbf{14.10} & \textbf{8.08} & \textbf{5.19} & \textbf{3.91} & \textbf{3.21} & \textbf{2.95} \\
\midrule
\multicolumn{10}{l}{CIFAR-10, Inception Score $\uparrow$} \\
\quad Gaussian & 1.13 & 2.92 & 6.17 & 8.11 & 8.67 & 8.98 & 9.17 & 9.28 & \textbf{9.40} \\
\quad GMM untrained & 3.77 & 5.46 & 7.59 & \textbf{8.42} & \textbf{8.86} & 9.00 & 9.18 & 9.27 & 9.37 \\
\quad GMM trained & \textbf{3.83} & \textbf{5.52} & \textbf{7.67} & \textbf{8.42} & 8.81 & \textbf{9.07} & \textbf{9.20} & \textbf{9.37} & 9.35 \\
\bottomrule
\end{tabular}
\caption{FID and Inception Score across Euler step counts, $10$k samples on
CelebA-HQ $256$ ($K = 1024$) and $50$k on CIFAR-10 ($K = 4096$), against the
full training sets.}
\label{tab:fid}
\end{table}

\begin{figure}[h]
\centering
\includegraphics[width=0.85\textwidth]{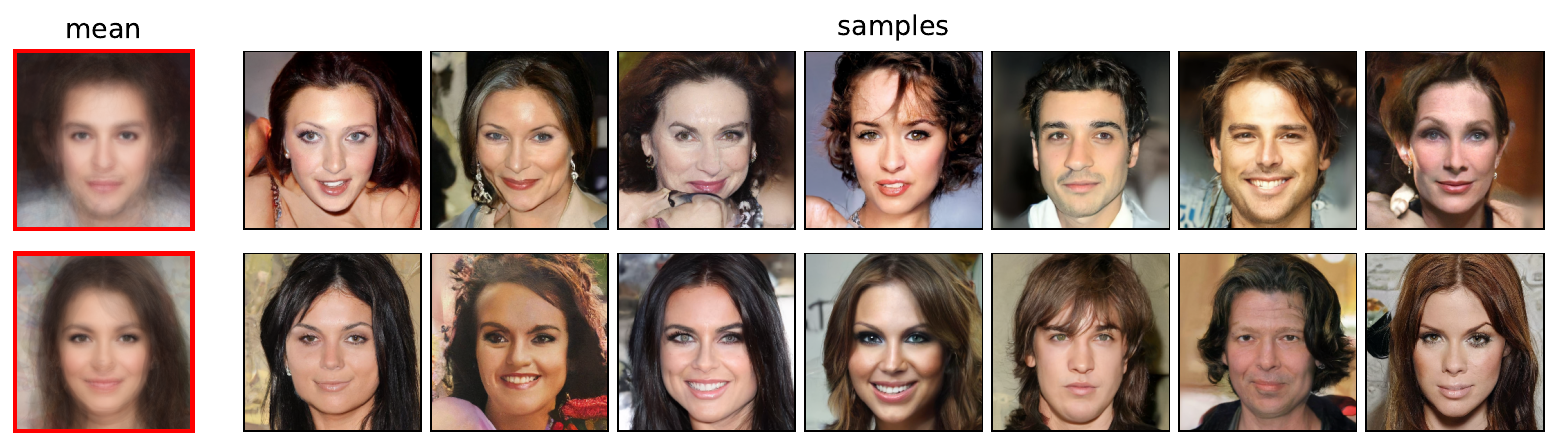}
\caption{The mean of a single component (left) and seven samples of the
coupled model conditioned on it, $256$ Euler steps, on CelebA-HQ. The
CIFAR-10 counterpart is in Appendix~\ref{app:additional}.}
\label{fig:components}
\end{figure}

Figure~\ref{fig:components} shows samples conditioned on a given component. The
samples inherit the palette and rough composition of the component mean and
nothing more; identity, class and background vary freely within the slice.

\subsection{The model does not memorize the training set}\label{sec:memorization}

\begin{figure}[h]
\centering
\includegraphics[width=0.85\textwidth]{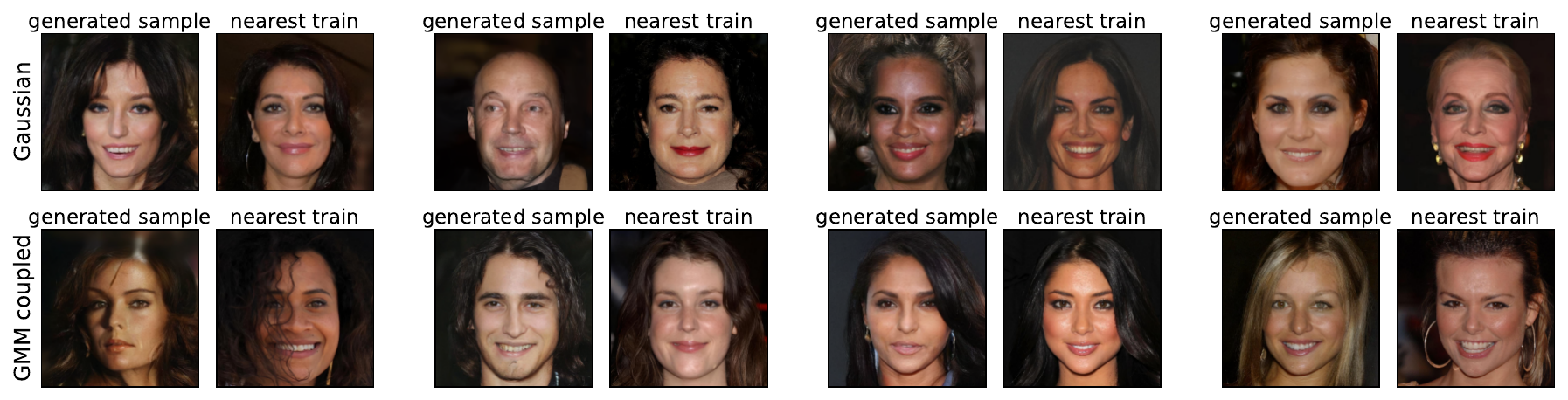}
\caption{The four generated samples closest to the training set, out of
$1000$, in pixel space, each next to its nearest training image, on CelebA-HQ.
The CIFAR-10 counterpart is in Appendix~\ref{app:additional}.}
\label{fig:memorization}
\end{figure}

Since the source is fitted to the training images, a natural worry is that
the model overfits the training set.
We draw $1000$ samples from our trained coupled models and $1000$ from the
Gaussian models, all with $256$-step Euler. We record the pixel-space distance between
every sample and its nearest training image and report the statistics in Table~\ref{tab:memorization}; the statistics in the Inception
latent space, which agree, are in Appendix~\ref{app:additional}. We also show
the closest pairs in Figure~\ref{fig:memorization}.

On both datasets and in both spaces, the coupled and Gaussian
distributions are very similar at every reported statistic. On CIFAR-10, the
statistics also show that both are similar to the distance distribution
between the training and test sets. The coupling does not move samples toward
the training set.

\begin{table}[h]
\centering
\small
\begin{tabular}{lccccccc}
\toprule
 & mean & std & min & $p_{1}$ & $p_{5}$ & median & max \\
\midrule
\multicolumn{8}{l}{CelebA-HQ, pixel $L^2$} \\
\quad GMM coupled & 162.0 & 22.9 & 95.8 & 110.9 & 125.5 & 161.8 & 231.4 \\
\quad Gaussian & 159.7 & 22.7 & 95.3 & 111.2 & 123.9 & 159.2 & 241.3 \\
\midrule
\multicolumn{8}{l}{CIFAR-10, pixel $L^2$} \\
\quad GMM coupled & 18.35 & 3.98 & 6.00 & 9.12 & 11.66 & 18.51 & 30.92 \\
\quad Gaussian & 18.10 & 4.24 & 6.69 & 8.39 & 11.24 & 18.15 & 30.09 \\
\quad test set & 18.11 & 4.37 & 0.51 & 8.02 & 10.67 & 18.21 & 36.10 \\
\bottomrule
\end{tabular}
\caption{Distance of $1000$ samples per model to their nearest training
image, in pixel space.}
\label{tab:memorization}
\end{table}

\section{Conclusion}

We replaced the Gaussian source of flow matching with a fitted Gaussian mixture, coupled to the data through  an optimal entropic coupling. A pretrained velocity network is reused without retraining and the target distribution is left intact. The coupling adds a correction that is mostly closed-form, and the closed-form part alone already delivers most of the improvement; a small correction network learns the rest. On two published checkpoints, the coupled model improves FID at every step count, reaches the quality of the baseline on CelebA-HQ with half the steps, and does not memorize the training set. Since the base network stays untouched, the source becomes a design parameter: a different mixture costs a mixture fit and a correction network, not a full retraining. Combining the coupling with accelerations that act on the velocity itself, such as distillation or reflow, and moving the construction to the latent space of larger models, are natural next steps.

\section*{Acknowledgment}
The research leading to these results was supported by the European Research Council
(ERC) under European Union’s Horizon 2020 (H2020), Grant Agreement - Project No
101020573 FunLearn and by the Swiss National Science Foundation (Grant
200020\_219356).

\section*{Reproducibility Statement}
All experiments use the two publicly released Rectified Flow checkpoints of \citet{liu2023rf}, the DDPM++ model for CIFAR-10 and the NCSN++ model for CelebA-HQ $256$, with their released configurations and no change to their weights. The datasets, their preprocessing and the flip-symmetrization of the CIFAR-10 training set are described in Section~\ref{sec:image-setup}. The mixture fit is described in Section~\ref{sec:experiments} and Appendix~\ref{app:recipe}, and the selection of $K$ and the fitted scales in Section~\ref{sec:image-setup}; the architecture of the correction network is given in Appendix~\ref{app:details} and its full training recipe, together with the evaluation protocol and the parity check against the published FID, in Appendix~\ref{app:recipe}. The two-dimensional experiment of Section~\ref{sec:toy} is fully specified there and runs on a CPU in minutes. The code for fitting the mixtures, training the correction networks and sampling, with the configuration of every table, is available at \url{https://github.com/CoupledGMMFlow/coupled-gmm-flow}.

\section*{The Use of Large Language Models}
Large language models were used as an assistant in four ways. To aid and polish the writing, by rephrasing sentences. For retrieval and discovery, to find and check related work. For research execution, to help write and document the evaluation, sampling and orchestration code, and to launch and monitor the evaluation runs under the authors' instructions. For the mathematical claims, to assist with derivations and to check the proofs, which the authors verified.

\bibliographystyle{iclr2027_conference}
\bibliography{refs}

\clearpage
\appendix

\section{Proofs}
\label{app:proofs}

\subsection{Proof of the score-velocity identity \eqref{eq:centered-tools}}
\label{app:centered}

The marginal of $\mat Z_t = t \mat X_1 + (1-t)\eps$ is $\rho_t(\z) = \E_{\mat X_1 \sim p_1}\big[\gN(\z;\, t \mat X_1,\, (1-t)^2\Id)\big]$, and by \eqref{eq:velocity-def} the velocity field is $\mat u_t(\z) = \E[\mat X_1 - \eps \mid \mat Z_t = \z]$. Next, write
\begin{equation}
    \z = \E[\mat Z_t \mid \mat Z_t = \z] = t\E[\mat X_1 \mid \mat Z_t = \z] + (1-t)\E[\eps \mid \mat Z_t = \z],
\end{equation}
which is equivalent to
\begin{equation}
    \E[\eps \mid \mat Z_t = \z] = \frac{1}{1-t} \Big( \z - \E[t \mat X_1 \mid \mat Z_t = \z] \Big).
\end{equation}
We also recall Tweedie's formula, which in our setting reads
\begin{equation}
    \E[t \mat X_1 \mid \mat Z_t = \z] = \z + (1-t)^2\nabla \log \rho_t(\z).
\end{equation}
The formulas above imply that
\begin{align}
    \mat u_t(\z) &= \E[\mat X_1 \mid \mat Z_t = \z] - \E[\eps \mid \mat Z_t = \z] \nonumber\\
    &= \frac{1}{t}\E[t \mat X_1 \mid \mat Z_t = \z] - \frac{1}{1-t} \Big( \z - \E[t \mat X_1 \mid \mat Z_t = \z] \Big) \nonumber\\
    &= \frac{1}{t}\Big( \z + (1-t)\, \nabla \log \rho_t(\z) \Big),
\end{align}
which is the desired result. \qed

\subsection{Derivation of the guidance identity \eqref{eq:tilt}}
\label{app:tilt-derivation}

The slice marginal $\rho^{(k)}_t$ is the conditional density of $\mat Z_t$ given $\kappa = k$, so the law of total probability gives $\sum_{k=1}^{K} \pi_k \rho^{(k)}_t = \rho_t$, and Bayes' rule implies that
\begin{equation}
\label{eq:slice-marginal}
    \rho^{(k)}_t(\z) = \frac{q_t(k \mid \z)}{\pi_k}\, \rho_t(\z) .
\end{equation}
The slice velocities are now related to the marginal one in closed form. Taking logarithms in \eqref{eq:slice-marginal} and differentiating in $\z$, the constant $\log \pi_k$ drops and
\begin{equation}
\label{eq:score-split}
    \nabla \log \rho^{(k)}_t(\z) = \nabla \log \rho_t(\z) + \nabla_{\z} \log q_t(k \mid \z) .
\end{equation}
The slice problem is itself a standard Gaussian-source problem, so \eqref{eq:centered-tools} applies to it and yields \eqref{eq:tilt}. \qed

\subsection{Proof of Proposition~\ref{prop:optimal-coupling}}
\label{app:proof-coupling}

We have that $\E_{\mat X_1 \sim p_1} \KL\big( \Gamma(\cdot \mid \mat X_1) \,\|\,
\Cat(\pib) \big) = \KL( \Gamma \,\|\,  \Cat(\pib) \otimes p_1)$. Therefore, Problem \eqref{eq:entropic-ot} is equivalent to the standard entropic optimal transport (EOT) problem
\begin{equation}
\label{pb:EOTstandard}
    \min_{\Gamma \in \Pi(\Cat(\pib), p_1)}\; \E_{(\kappa, \mat X_1) \sim \Gamma} \| \mat X_1 - \mub_\kappa\|^2 \;+\; \eta\, \KL( \Gamma \,\|\,  \Cat(\pib) \otimes p_1).
\end{equation}
Under the assumption that $\mathrm{Cat}(\pib)$ and $p_1$ have finite second-order moments (this holds automatically if $p_1$ is compactly supported), which is the case here, the existence and uniqueness of the solution to Problem~\eqref{pb:EOTstandard} is a standard result \citep{chewiStatisticalOptimalTransport2024}. Moreover, with respect to the product of the counting and the Lebesgue measure, the optimal solution has density
\begin{equation}
\label{eq:optimaldensityEOT}
    \Gamma(k, \x_1) = \pi_k p_1(\x_1) e^{(\varphi(\x_1) + \psi_k - \|\x_1 - \mub_k\|^2)/\eta},
\end{equation}
where $\varphi(\x_1)$ and $\psi_k$ are dual variables \citep[Proposition 4.3]{chewiStatisticalOptimalTransport2024}, also referred to as \emph{EOT potentials}. They satisfy the Schr{\"o}dinger equation \citep[Section 4.2]{chewiStatisticalOptimalTransport2024}
\begin{equation}
\label{eq:schrodingerphi}
    \varphi(\x_1) = -\eta \log \sum_{i=1}^{K} \pi_i e^{(\psi_i - \|\x_1 - \mub_i\|^2)/\eta}.
\end{equation}
Plugging \eqref{eq:schrodingerphi} into \eqref{eq:optimaldensityEOT} yields the closed form expression
\begin{align}
    \Gamma(k \mid \x_1) = \frac{\Gamma(k,\x_1)}{p_1(\x_1)} &= \pi_k e^{(\varphi(\x_1) + \psi_k - \|\x_1 - \mub_k\|^2)/\eta} \nonumber\\
    &= \frac{\pi_k e^{(\psi_k - \|\x_1 - \mub_k\|^2)/\eta}}{\sum_{i=1}^{K} \pi_i e^{(\psi_i - \|\x_1 - \mub_i\|^2)/\eta}} \nonumber\\
    &= \softmax_k\Big( \log \pi_k + \frac{\psi_k - \|\x_1 - \mub_k\|^2}{\eta} \Big).
\end{align}

The dual variable $\psi_k$ corresponds to the Lagrange multiplier of the $\mathrm{Cat}(\pib)$-marginal constraint
\begin{equation}
\label{eq:cat-pi-marginal}
    \int_{\R^d} \Gamma(k, \x_1) \, \mathrm{d}\x_1 = \pi_k \iff \E_{\mat X_1 \sim p_1} [ \Gamma(k \mid \mat X_1)] = \pi_k
\end{equation}
for all $k = 1, \ldots, K$.

Next, set $\eta = 2\tau^2$ and $\vec \psi = \vec 0$. Then $\Gamma(k \mid \x_1) = p_0(k \mid \x_1)$, the mixture posterior~\eqref{eq:responsibility}. Its $p_1$-marginal is automatic, and Condition~\eqref{eq:weight-condition} is exactly the remaining $\Cat(\pib)$-marginal constraint~\eqref{eq:cat-pi-marginal}. Thus, under~\eqref{eq:weight-condition}, the choice $\vec \psi = \vec 0$ satisfies the marginal constraint that determines the optimal coupling. Consequently, the posterior responsibility is the entropic optimal coupling, and uniqueness of the solution to~\eqref{pb:EOTstandard} shows that it is the unique optimum. \qed

\subsection{Proof of Proposition~\ref{prop:stationarity}}
\label{app:proof-stationarity}

Since $p_0(k \mid \x_1) = \pi_k\, \mathcal{N}(\x_1; \mub_k, \tau^2\mathbf{I}) / p_0(\x_1)$, differentiating the log-likelihood with respect to $\pi_k$ gives
\begin{equation}
    \frac{\partial}{\partial \pi_k}\, \E_{p_1} \big[ \log p_0(\mat X_1) \big]
    = \E_{p_1}\!\left[ \frac{\gN(\mat X_1; \mub_k, \tau^2\Id)}{p_0(\mat X_1)} \right]
    = \frac{\E_{p_1}\big[\, p_0(k \mid \mat X_1) \,\big]}{\pi_k}.
\end{equation}
We optimize over the open simplex, so there are no inequality constraints to include in the Lagrangian, we only introduce a multiplier for the normalization constraint. The Lagrangian is
\begin{equation}
    \mathcal{L}(\pib, \lambda) = \E_{p_1} \big[ \log p_0(\mat X_1) \big] - \lambda \bigg(\sum_{k=1}^{K} \pi_k - 1 \bigg),
\end{equation}
with $\lambda \in \R$. Its stationary points satisfy
\begin{equation}
    \frac{\E_{p_1}\big[\, p_0(k \mid \mat X_1) \,\big]}{\pi_k} - \lambda = 0
\end{equation}
for all $k=1, \ldots, K$, or equivalently,
\begin{equation}
\label{eq:pi-stationary}
    \E_{p_1}\big[\, p_0(k \mid \mat X_1) \,\big] = \lambda \pi_k, \quad k=1,\ldots, K.
\end{equation}
Summing \eqref{eq:pi-stationary} over $k$ and using
$\sum_{k=1}^{K} p_0(k \mid \x_1) = 1$ and $\sum_{k=1}^{K} \pi_k = 1$ gives $\lambda = 1$, which yields Condition \eqref{eq:weight-condition}. \qed

\subsection{Proof of Proposition~\ref{prop:marginal}}
\label{app:proof-marginal}

Conditioned on $\kappa = k$, the construction of Section~\ref{sec:guidance} given by
\begin{equation}
\label{eq:X0-given-kappa}
    \mat X_t = t \mat X_1 + (1-t) \mat X_0, \quad \mat X_0 = \mub_k + \tau \eps, \quad \eps \sim \gN(\mathbf{0},\Id)
\end{equation}
is standard flow matching between $\mathcal{N}(\mub_k, \tau^2\mathbf{I})$ and the slice $p_1^{(k)}$. By the marginal-generation property of the exact velocity field \citep{lipman2023, albergo2023}, integrating the corresponding exact conditional velocity field
\begin{equation}
    \mat v_t(\x \mid k) = \E[\mat X_1 - \mat X_0 \mid \mat X_t = \x, \kappa=k]
\end{equation}
from $t=0$ to $t = 1$ with initial point $\x_0 \sim \gN(\mub_k, \tau^2\Id)$ produces $\x_1 \sim p_1^{(k)}$. Mixing over $\kappa \sim \Cat(\pib)$ gives $\sum_{k=1}^{K} \pi_k p_1^{(k)} = p_1$ by the definition of the slices~\eqref{eq:slices}. \qed

An analytical expression of $\mat v_t(\x \mid k)$ is computed in Proposition~\ref{prop:reparam}.

\subsection{Proof of Proposition~\ref{prop:reparam}}
\label{app:proof-reparam}

Conditioned on $\kappa = k$, we have that  $\mat X_t = t \mat X_1 + (1-t)\mub_k + (1-t)\tau\,\eps$. With $c(t) = t + (1-t)\tau$ and $s(t) = t/c(t)$,
\begin{align}
    c(t) \mat Z_{s(t)} + (1-t)\mub_k &= c(t)\Big[ \tfrac{t}{c(t)} \mat X_1 + \big(1 - \tfrac{t}{c(t)}\big)\eps \Big] + (1-t)\mub_k \nonumber\\
    &= t \mat X_1 + (1-t)\tau\eps + (1-t)\mub_k \nonumber\\
    &= \mat X_t \mid \{\kappa = k\} ,
\end{align}
which is \eqref{eq:shift}. Next, we compute the conditional velocity
\begin{equation}
    \mat v_t(\x \mid k) = \E[\mat X_1 - \mat X_0 \mid \mat X_t = \x, \kappa = k] = \E[\mat X_1 - \mub_k - \tau\eps \mid \mat X_t = \x, \kappa=k],
\end{equation}
due to~\eqref{eq:X0-given-kappa}. We can verify that
\begin{equation}
    \mat X_1 - \mub_k - \tau\eps = (1-\tau) \mat Z_{s(t)} + \frac{\tau}{c(t)}(\mat X_1 - \eps) - \mub_k,
\end{equation}
so that
\begin{equation}
    \mat v_t(\x \mid k) = (1-\tau)\E[ \mat Z_{s(t)} \mid \mat X_t = \x, \kappa = k] + \frac{\tau}{c(t)} \E[\mat X_1 - \eps \mid \mat X_t = \x, \kappa = k] - \mub_k.
\end{equation}
By~\eqref{eq:shift}, given $\kappa=k$, $\mat X_t$ is a deterministic and invertible function of $\mat Z_{s(t)}$, so that the events
\begin{equation}
    \{\mat X_t = \x \mid \kappa = k\} = \{\mat Z_{s(t)} = \z \mid \kappa=k\}, \quad \z = \frac{\x - (1-t)\mub_k}{c(t)}
\end{equation}
are equal. Therefore,
\begin{align}
\label{eq:developpementCondVelocity}
    \mat v_t(\x \mid k) &= (1-\tau)\z + \frac{\tau}{c(t)} \E[\mat X_1 - \eps \mid \mat Z_{s(t)}=\z, \kappa = k] - \mub_k \nonumber\\
    &= (1-\tau)\z + \frac{\tau}{c(t)} \mat u_{s(t)}^{(k)}(\z) - \mub_k \nonumber\\
    &= \frac{1-\tau}{c(t)}\big(\x - (1-t)\mub_k\big) + \frac{\tau}{c(t)} \mat u^{(k)}_{s(t)}\!\left(\frac{\x - (1-t)\mub_k}{c(t)}\right) - \mub_k.
\end{align}
The mixture form of the marginal velocity is the law of total probabilities over $\kappa$. \qed

\subsection{Proof of Proposition~\ref{prop:exact-corr}}
\label{app:proof-exact-corr}

Since $\mat Z_t$ depends on $\kappa$ only through $\mat X_1$, the chain
$\kappa \to \mat X_1 \to \mat Z_t$ is Markov, so conditioning on
$\mat X_1$ inside the conditional expectation and using the tower property,
\begin{equation}
\label{eq:q-tower}
    q_t(k \mid \z) = \E\big[ \Gamma(k \mid \mat X_1) \mid \mat Z_t = \z \big] = \E\big[p_0(k \mid \mat X_1) \mid \mat Z_t = \z\big].
\end{equation}
Therefore,
\begin{equation}
\label{eq:q-integral}
    q_t(k \mid \z) = \int_{\R^d} p_0(k \mid \x_1) p(\x_1 \mid \z)\, \mathrm{d}\x_1.
\end{equation}
The first factor is a GMM posterior, given by
\begin{equation}
    p_0(k \mid \x_1) = \frac{\pi_k\, \gN(\x_1; \mub_k, \tau^2 \Id)}{p_0(\x_1)}.
\end{equation}
Recalling that $\rho_t$ is the density of $\mat Z_t$, the second factor is
\begin{equation}
    p(\x_1 \mid \z) = \frac{p_{\mat Z_t \mid \mat X_1 = \x_1}(\z) p_1(\x_1)}{\rho_t(\z)} = \frac{p_1(\x_1)\, \gN(\z; t\x_1, (1-t)^2 \Id)}{\rho_t(\z)},
\end{equation}
due to the fact that $\mat Z_t \mid \{\mat X_1 = \x_1\} = t\x_1 + (1-t)\eps$ with $\eps \sim \gN(\mat 0, \Id)$. We then compute that
\begin{equation}
\label{eq:conjugacy}
    \gN(\x_1; \mub_k, \tau^2 \Id) \, \gN(\z; t\x_1, (1-t)^2 \Id) = \gN(\z;\, t\mub_k,\, \sigma_t^2 \Id) \, \gN\big(\x_1;\, \mat m_k(\z,t),\, \nu_t \Id\big) ,
\end{equation}
with $\sigma_t^2$, $\mat m_k(\z, t)$ and $\nu_t$ as in~\eqref{eq:mk}. Therefore,
\begin{align}
    q_t(k \mid \z) &= \frac{\pi_k \gN(\z;\, t\mub_k,\, \sigma_t^2 \Id)}{\rho_t(\z)} \int_{\R^d} \frac{p_1(\x_1)}{p_0(\x_1)} \gN\big(\x_1;\, \mat m_k(\z,t),\, \nu_t \Id\big) \, \mathrm{d}\x_1 \nonumber\\
    &= \frac{\pi_k \gN(\z;\, t\mub_k,\, \sigma_t^2 \Id)}{\rho_t(\z)} \E_{\mat X \sim \gN(\mat m_k(\z,t),\, \nu_t \Id)} \! \left[ \frac{p_1(\mat X)}{p_0(\mat X)} \right] \nonumber\\
    &= \frac{\pi_k \gN(\z;\, t\mub_k,\, \sigma_t^2 \Id)}{\rho_t(\z)} F_t\big(\mat m_k(\z, t)\big),
\end{align}
with $F_t$ as in~\eqref{eq:mk}. Since $\sum_{k=1}^{K} q_t(k \mid \z) = 1$, we have that
\begin{equation}
    \rho_t(\z) = \sum_{k=1}^{K} \pi_k \gN(\z;\, t\mub_k,\, \sigma_t^2 \Id) F_t(\mat m_k(\z, t)),
\end{equation}
so that
\begin{align}
    q_t(k \mid \z) = \frac{\pi_k \gN(\z;\, t\mub_k,\, \sigma_t^2 \Id) F_t\big(\mat m_k(\z, t)\big) }{\sum_{i=1}^{K} \pi_i \gN(\z;\, t\mub_i,\, \sigma_t^2 \Id) F_t\big(\mat m_i(\z, t)\big)} &= \frac{\pi_k \mathrm{exp}\Big(-\frac{\|\z - t\mub_k\|^2}{2\sigma_t^2}\Big) F_t\big(\mat m_k(\z, t)\big) }{\sum_{i=1}^{K} \pi_i \mathrm{exp}\Big(-\frac{\|\z - t\mub_i\|^2}{2\sigma_t^2}\Big) F_t\big(\mat m_i(\z, t)\big)} \nonumber\\
    &= \softmax_k \ell_k(\z, t)
\end{align}
with $\ell_k(\z, t)$ as in~\eqref{eq:logits}.

For the correction, recall that by~\eqref{eq:tilt},
\begin{equation}
    \mat u^{(k)}_t(\z) - \mat u_t(\z) = \frac{1-t}{t} \nabla_{\z} \log q_t(k \mid \z).
\end{equation}
We compute
\begin{equation}
\label{eq:softmax-grad}
    \nabla_{\z} \log q_t(k \mid \z) = \nabla_{\z} \ell_k(\z, t) - \sum_{j=1}^{K} q_t(j \mid \z)\, \nabla_{\z} \ell_j(\z, t),
\end{equation}
and
\begin{equation}
\label{eq:grad-lz}
    \nabla_{\z} \ell_k(\z, t) = \frac{t\mub_k - \z}{\sigma_t^2} + \frac{t\tau^2}{\sigma_t^2} [\nabla \log F_t](\mat m_k(\z, t)).
\end{equation}
Plugging~\eqref{eq:grad-lz} into~\eqref{eq:softmax-grad} and multiplying by the prefactor $\frac{1-t}{t}$ of~\eqref{eq:tilt} yields~\eqref{eq:corr-exact}. \qed

\section{Experimental Details}

\subsection{Architecture of the Correction Network}
\label{app:details}

The network $\mat W_{\phi}$ maps its six input channels, the concatenation of
$\mat m_k$ and $\z$, through a $3 \times 3$ convolution to $64$
channels. The result is average-pooled by a factor of $2$ and passed through
four residual blocks that run at the pooled resolution; their output is
upsampled back by nearest-neighbor interpolation and summed with the
full-resolution activation of the input convolution, a skip that restores the
detail the pooling discards while making every block four times cheaper. A
group normalization, a SiLU, and a $3 \times 3$ convolution back to $3$
channels produce the output.

Each residual block is conditioned in the style of adaptive layer
normalization \citep{peebles2023dit}: the block contains two convolutions,
each preceded by a group normalization without learned affine parameters,
and a linear map of the conditioning vector produces a per-channel scale and
shift for both normalizations. The conditioning vector has $16$ dimensions
and is the concatenation of two halves. The time half is a sinusoidal
embedding of $t$ passed through a two-layer MLP. The posterior half is
$q^{\mathrm{ref}}_t(\cdot \mid \z)$ passed through a two-layer MLP; the
posterior is the only input whose dimension depends on $K$, so this
projection is the only $K$-dependent part of $\mat W_{\phi}$. The probabilities are
fed rather than the logits, a linear map of a probability vector being a
convex combination of $K$ learned vectors and therefore bounded at every $t$,
while the spread of the logits grows without bound as the posterior
sharpens. The output convolution is initialized at zero, so at initialization
$\mat W_{\phi} = \mathbf{0}$ and the model starts exactly as the untrained model
of Section~\ref{sec:parameterization}.

\subsection{Training Recipes and Evaluation Protocol}
\label{app:recipe}

The mixtures are fitted with Adam at learning rate $10^{-2}$ for $2000$
full-batch steps without decay. The means are initialized at random, the
weights uniformly, and the scale $\tau$ at a value far above the spread of
the data, about $3 \cdot 10^{3}$. At that scale every responsibility is
uniform, so every component receives the full gradient of the likelihood
from the first step and none collapses to zero weight, a common failure of
mixture fits with many components. As the means move onto the data, $\tau$
decreases by itself to its final value.

The correction network $\mat W_{\phi}$ is trained by minimizing \eqref{eq:head-loss} with
Adam at learning rate $2 \cdot 10^{-4}$, batch size $16$, $100$k steps with a
$1$k-step linear warmup and cosine decay to zero, gradient clipping at $1$,
$t$ uniform on the release's range $[10^{-3}, 1]$, and an EMA of
$\mat W_{\phi}$ with decay $0.9999$, which is what sampling deploys.

Before the evaluations of Section~\ref{sec:fewstep}, we verified that our
pipeline reproduces the published numbers. With the
release's own adaptive RK45 solver \citep[Dormand--Prince, tolerances
$10^{-5}$]{dormand1980rk45} we obtain FID $2.82$ on CIFAR-10 against a
published $2.58$, the residual gap being the known discrepancy between TF-hub
and torch Inception features \citep{parmar2022cleanfid}.

\subsection{Additional Results}
\label{app:additional}

\begin{figure}[h]
\centering
\includegraphics[width=\textwidth]{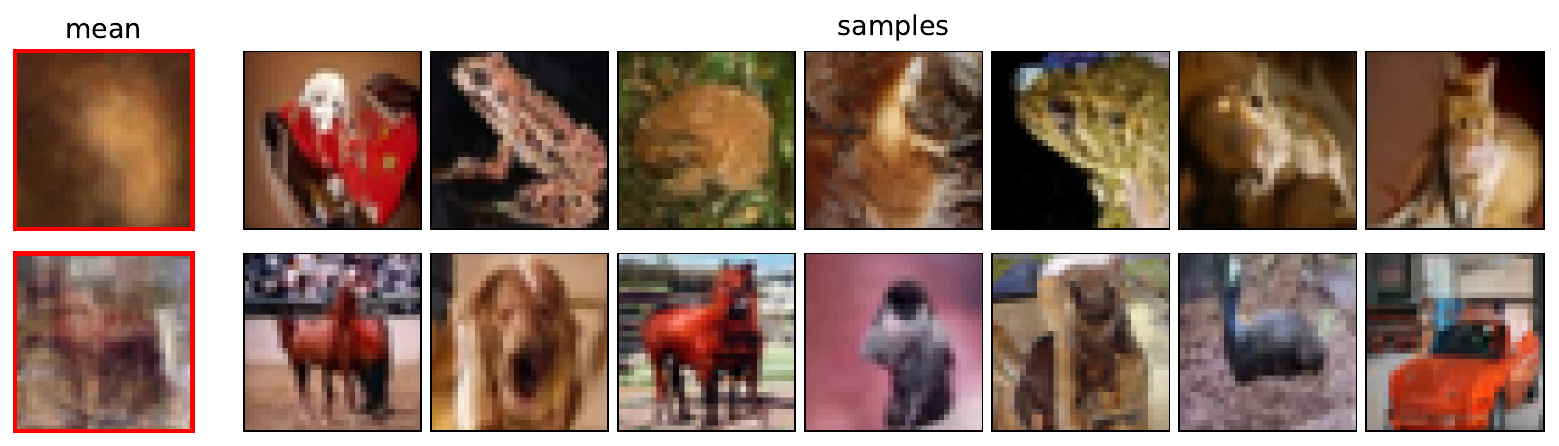}
\caption{CIFAR-10: the mean of a single component (left) and seven samples
of the coupled model conditioned on it, $256$ Euler steps.}
\label{fig:components-cifar}
\end{figure}

\begin{figure}[h]
\centering
\includegraphics[width=\textwidth]{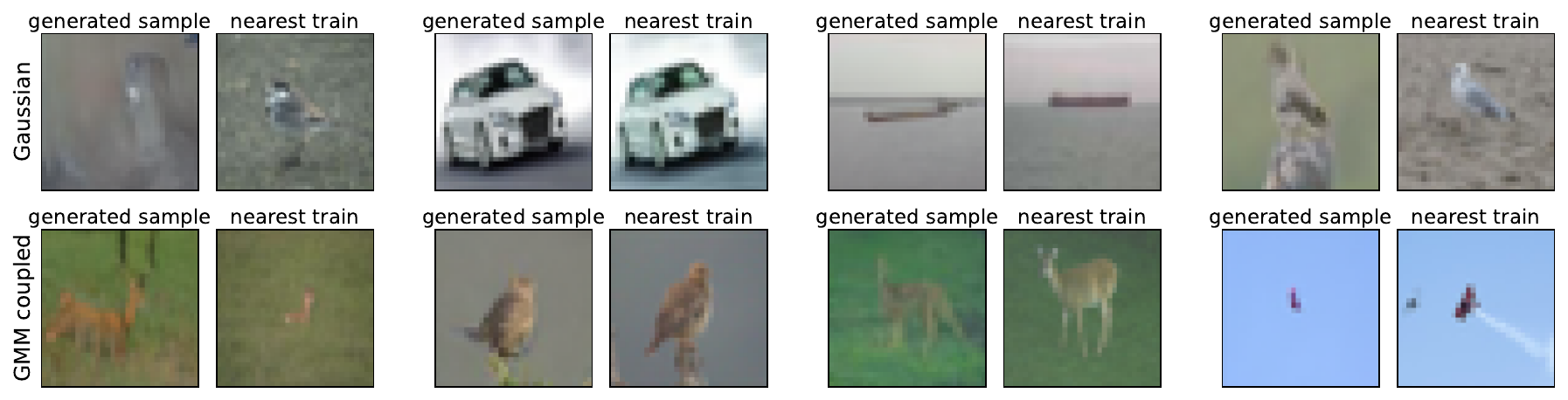}
\caption{CIFAR-10: the four generated samples closest to the flip-symmetrized
training set, out of $1000$, by pixel $L^2$, each next to its nearest
training image.}
\label{fig:memorization-cifar}
\end{figure}

Tables~\ref{tab:ksweep-fid-celeba} and~\ref{tab:ksweep-fid-cifar} report
the FID of the untrained model of Section~\ref{sec:parameterization} at
every mixture size $K$ of the sweep of Section~\ref{sec:experiments}, each
with its own maximum-likelihood mixture and the same sampler as
Table~\ref{tab:fid}; the trained model is reported in Table~\ref{tab:fid}
at the selected $K$. On CelebA-HQ, the FID improves with $K$ at every step count up
to $64$ steps, and every mixture beats the Gaussian at every step count. On
CIFAR-10, the FID improves with $K$ up to $4$ steps, with the exception of the
single-step FID of the smallest mixture, which is worse than the Gaussian one. On both datasets, the differences between the mixture sizes
shrink as the step count grows, which is what
Proposition~\ref{prop:marginal} predicts since every mixture targets
the same marginal.
Table~\ref{tab:memorization-latent} reports the nearest-neighbor statistics
of Section~\ref{sec:memorization} in the Inception latent space, and
Figures~\ref{fig:components-cifar} and~\ref{fig:memorization-cifar} the
CIFAR-10 counterparts of Figures~\ref{fig:components}
and~\ref{fig:memorization}.

\begin{table}[h]
\centering
\small
\setlength{\tabcolsep}{3pt}
\begin{tabular}{lccccccccc}
\toprule
Euler steps & 1 & 2 & 4 & 8 & 16 & 32 & 64 & 128 & 256 \\
\midrule
\quad Gaussian & 258.40 & 182.43 & 137.16 & 89.01 & 48.53 & 25.08 & 15.00 & 11.45 & 10.01 \\
\quad GMM untrained, $K=4$ & 252.49 & 174.30 & 122.69 & 83.35 & 48.33 & 24.38 & 14.24 & 10.99 & 9.80 \\
\quad GMM untrained, $K=16$ & 242.64 & 163.23 & 111.51 & 75.50 & 44.80 & 22.94 & 13.97 & 10.54 & 9.77 \\
\quad GMM untrained, $K=64$ & 233.41 & 153.49 & 106.58 & 71.06 & 41.19 & 21.54 & 13.59 & 10.60 & 9.66 \\
\quad GMM untrained, $K=256$ & 218.41 & 141.49 & 99.99 & 66.99 & 38.61 & 20.94 & 13.01 & \textbf{10.42} & 9.73 \\
\quad GMM untrained, $K=1024$ & \textbf{186.94} & \textbf{123.86} & \textbf{93.61} & \textbf{63.60} & \textbf{37.12} & \textbf{19.91} & \textbf{12.81} & 10.63 & \textbf{9.63} \\
\bottomrule
\end{tabular}
\caption{CelebA-HQ $256$, FID $\downarrow$ of the untrained model
($\mat W_{\phi} = \mathbf{0}$) for every mixture size $K$, $10$k samples
against the full training set, uniform Euler grid in $s(t)$. The $K = 1024$
row is the one of Table~\ref{tab:fid}.}
\label{tab:ksweep-fid-celeba}
\end{table}

\begin{table}[h]
\centering
\small
\setlength{\tabcolsep}{3pt}
\begin{tabular}{lccccccccc}
\toprule
Euler steps & 1 & 2 & 4 & 8 & 16 & 32 & 64 & 128 & 256 \\
\midrule
\quad Gaussian & 371.25 & 173.07 & 52.95 & 17.20 & 8.59 & \textbf{5.47} & 4.13 & 3.43 & 3.05 \\
\quad GMM untrained, $K=4$ & 415.10 & 142.93 & 43.32 & 15.61 & \textbf{8.27} & 5.50 & \textbf{4.01} & 3.34 & 3.02 \\
\quad GMM untrained, $K=16$ & 357.96 & 130.65 & 39.43 & 15.12 & 8.40 & 5.51 & 4.04 & 3.40 & 3.03 \\
\quad GMM untrained, $K=64$ & 312.40 & 120.00 & 36.35 & 14.81 & 8.39 & 5.57 & 4.02 & 3.38 & 2.98 \\
\quad GMM untrained, $K=256$ & 250.91 & 108.81 & 34.59 & 14.89 & 8.47 & 5.54 & 4.05 & 3.38 & 2.97 \\
\quad GMM untrained, $K=1024$ & 197.89 & 95.62 & 32.97 & 14.59 & 8.49 & 5.58 & 4.05 & 3.34 & \textbf{2.95} \\
\quad GMM untrained, $K=4096$ & \textbf{151.43} & \textbf{85.62} & \textbf{31.36} & \textbf{14.49} & 8.48 & 5.65 & 4.13 & \textbf{3.32} & 2.98 \\
\bottomrule
\end{tabular}
\caption{CIFAR-10, FID $\downarrow$ of the untrained model
($\mat W_{\phi} = \mathbf{0}$) for every mixture size $K$, $50$k samples
against the full training set, uniform Euler grid in $s(t)$. The $K = 4096$
row is the one of Table~\ref{tab:fid}.}
\label{tab:ksweep-fid-cifar}
\end{table}

\begin{table}[h]
\centering
\small
\begin{tabular}{lccccccc}
\toprule
 & mean & std & min & $p_{1}$ & $p_{5}$ & median & max \\
\midrule
\multicolumn{8}{l}{CelebA-HQ, latent $L^2$} \\
\quad GMM coupled & 9.18 & 1.95 & 5.12 & 5.92 & 6.56 & 8.87 & 17.23 \\
\quad Gaussian & 9.11 & 1.94 & 5.30 & 5.83 & 6.43 & 8.83 & 18.09 \\
\midrule
\multicolumn{8}{l}{CIFAR-10, latent $L^2$} \\
\quad GMM coupled & 12.09 & 2.51 & 5.59 & 6.48 & 7.91 & 12.33 & 18.24 \\
\quad Gaussian & 11.99 & 2.53 & 5.56 & 6.31 & 7.86 & 12.22 & 18.18 \\
\quad test set & 11.66 & 2.65 & 2.37 & 5.69 & 7.37 & 11.65 & 20.48 \\
\bottomrule
\end{tabular}
\caption{Distance of $1000$ samples per model to their nearest training
image in the Inception latent space. The CIFAR-10 test-set rows are a
reference of novel images.}
\label{tab:memorization-latent}
\end{table}

\end{document}